%% file: main.tex
\documentclass[journal]{IEEEtran}
\usepackage{epsfig}
\usepackage{graphicx}
\usepackage{amsmath}
\usepackage[pagebackref,breaklinks,colorlinks,citecolor=blue]{hyperref}
\usepackage{amssymb}
\usepackage{booktabs}
\usepackage{subfigure}
\usepackage{xcolor}
\usepackage{xspace}

\definecolor{DarkGreen}{rgb}{0.43, 0.68, 0.28}

\makeatletter
\DeclareRobustCommand\onedot{\futurelet\@let@token\@onedot}
\def\@onedot{\ifx\@let@token.\else.\null\fi\xspace}

\def\etal{\emph{et al}\onedot}

\ifCLASSINFOpdf
\else
\fi
\begin{document}
%

\title{ForgeryTTT: Zero-Shot Image Manipulation Localization with Test-Time Training}


%
%
%

\author{Weihuang Liu, Xi Shen, Chi-Man Pun,~\IEEEmembership{Senior Member,~IEEE},
        and~Xiaodong Cun
\thanks{

Weihuang Liu and Chi-Man Pun are with the Department of Computer and Information Science, Faculty of Science and Technology, University of Macau, Taipa, Macau (e-mail: nifangbaage@gmail.com; cmpun@umac.mo).

Xi Shen is with Intellindust (e-mail: shenxiluc@gmail.com).

Xiaodong Cun is with the School of Computing and Information Technology, Great Bay University, Dongguan, China (e-mail: vinthony@gmail.com).

Corresponding author: Chi-Man Pun and Xiaodong Cun.}
}

\newcommand{\lossmain}{\mathcal{L}_{\text{main}}}
\newcommand{\lossbce}{\mathcal{L}_{\text{bce}}}
\newcommand{\lossssl}{\mathcal{L}_{\text{ssl}}}
\newcommand{\lossdepth}{\mathcal{L}_{\text{depth}}}

\newcommand{\sslweight}{\lambda}
\newcommand{\encoder}{\mathcal{E}}

\newcommand{\enimagei}{\mathcal{E}_{k-1}}
\newcommand{\enimage}{\mathcal{E}}
\newcommand{\demain}{\mathcal{D}}
\newcommand{\deauxiliary}{\mathcal{C}}

\newcommand{\dropoutratio}{r}
\newcommand{\image}{i}
\newcommand{\mask}{m}
\newcommand{\imagetest}{i^{'}}
\newcommand{\masktest}{m^{'}}
\newcommand{\masktesti}{m^{'}_{k-1}}
\newcommand{\class}{c}

\maketitle


\input{submit/abs}


%
\IEEEpeerreviewmaketitle

\input{submit/intro}
\input{submit/related}
\input{submit/method}

\input{submit/exp}

\input{submit/conclusion}


%





\ifCLASSOPTIONcaptionsoff
  \newpage
\fi



%



\bibliographystyle{./Transactions-Bibliography/IEEEtran.bst}
\bibliography{IEEEabrv, submit/egbib}

%








\end{document}

%% file: submit/abs.tex

\begin{abstract}

Social media is increasingly plagued by realistic fake images, making it hard to trust content. Previous algorithms to detect these fakes often fail in new, real-world scenarios because they are trained on specific datasets. To address the problem, we introduce ForgeryTTT, the first method leveraging test-time training (TTT) to identify manipulated regions in images. The proposed approach fine-tunes the model for each individual test sample, improving its performance. ForgeryTTT first employs vision transformers as a shared image encoder to learn both classification and localization tasks simultaneously during the training-time training using a large synthetic dataset. Precisely, the localization head predicts a mask to highlight manipulated areas. Given such a mask, the input tokens can be divided into manipulated and genuine groups, which are then fed into the classification head to distinguish between manipulated and genuine parts. During test-time training, the predicted mask from the localization head is used for the classification head to update the image encoder for better adaptation. Additionally, using the classical dropout strategy in each token group significantly improves performance and efficiency. 
We test ForgeryTTT on five standard benchmarks. Despite its simplicity, ForgeryTTT achieves a 20.1$\%$ improvement in localization accuracy compared to other zero-shot methods and a 4.3$\%$ improvement over non-zero-shot techniques. 
Our code will be released upon publication.

\end{abstract}

%% file: submit/intro.tex
\section{Introduction}
\IEEEPARstart{T}{he}
rapid development of image editing techniques has made image manipulation much easier~\cite{cun2020improving,suvorov2022resolution,liu2023coordfill}. People can easily alter image content using editing tools, often fooling the human eye. Advanced post-processing techniques further reduce the detectability of fake images. However, the misuse of these techniques raises concerns about intentional image manipulation. Frequent digital image fraud incidents have led to serious doubts about the authenticity of digital images, diminishing social credibility. To address this issue, it is urgent to develop effective image manipulation localization algorithms.


Image manipulation often involves operations like splicing, copy-move, and removal, which leave distinct traces and artifacts. Image manipulation localization aims to identify these tampered regions. Traditional algorithms mainly rely on low-level clues such as noise~\cite{noi}, JPEG compression artifacts~\cite{lin2009fast}, and camera traces~\cite{cfa}. Deep learning-based approaches learn from large-scale datasets~\cite{mantra,hlstm,span}, demonstrating promising results.

\begin{figure}[t]
    \centering
    \includegraphics[width=\linewidth]{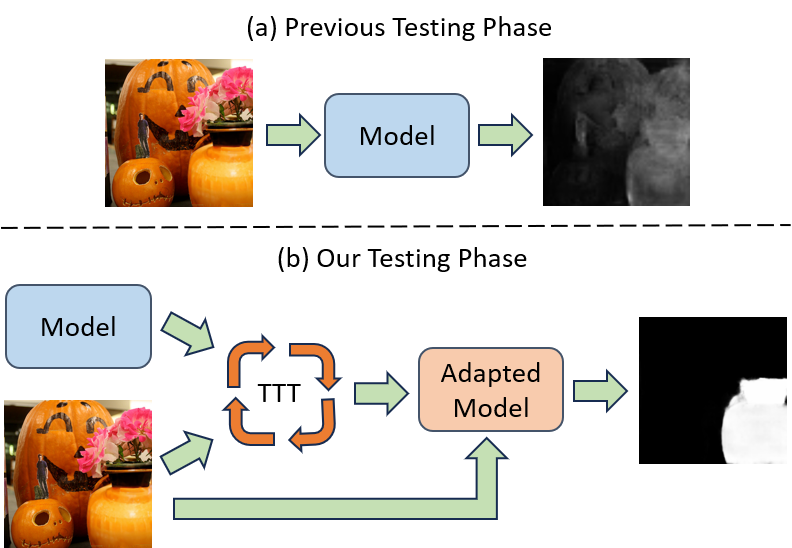}
    \caption{\textbf{Comparison of previous and our testing phase.} 
    Previous methods directly employ the models for forgery localization, while we first perform model adaptation for each image and then localize the forgery region.}
    \label{fig:brife_pipeline}
\end{figure}


While some progress has been made, existing approaches often fail in real-world scenarios. The rapid advancement of editing technologies, especially with generative artificial intelligence~\cite{hertz2022prompt,huang2024smartedit,qi2023fatezero}, makes it difficult to collect a comprehensive dataset of manipulated samples. As a result, current image manipulation localization algorithms face significant challenges in adapting to evolving forgery techniques.

In this work, we propose using test-time training (TTT)~\cite{schneider2020improving,wang2020tent,sun2020test} to address the issue. TTT adapts the pre-trained model to new targets during testing, showing excellent generalization ability by fine-tuning the model for each test sample. 
To the best of our knowledge, we are the first to explore TTT for image manipulation localization. As shown in Figure~\ref{fig:brife_pipeline}, our method is quite different from previous methods which directly employ the pre-trained model for inference. The proposed method first performs model adaptation for the given sample and then uses the adapted model for better prediction.

Typical auxiliary-head-based TTT methods use a well-designed auxiliary self-supervised objective function to update the model at test-time training~\cite{sun2020test,liu2021ttt++,gandelsman2022test}. Note that the effectiveness of TTT greatly depends on designing an auxiliary task closely related to the main task~\cite{liu2021ttt++}.
Our key motivation arises from the consistency between the coarse task of distinguishing whether the image is manipulated and the fine-grained task of predicting exact manipulation regions.
Furthermore, deep neural networks have showcased promising performance in image manipulation classification, despite the challenges presented by image manipulation localization. They can effortlessly determine whether the images have undergone manipulation.
This provides a favorable prerequisite and encourages us to investigate an effective auxiliary task based on image manipulation classification, which boosts the pre-trained image manipulation localization model at test time.

Specifically, our ForgeryTTT is built upon the commonly used encoder-decoder network for image manipulation localization.
It is composed of a shared image encoder of vision transformer~\cite{dosovitskiy2020image,liu2021swin} architecture, a localization head for forgery mask prediction, and a classification head for self-supervised image manipulation classification.
During training-time training, ForgeryTTT simultaneously learns image manipulation localization and self-supervised image manipulation classification. 
We follow the standard supervised protocol to train the localization branch. 
For the self-supervised classification head, we first divide the input tokens into manipulated and authentic groups via the given mask. Then we feed them into the classification head to distinguish between manipulated and authentic parts.
During test-time testing, the predicted mask from the localization head is used for the classification head to update the image encoder via the self-supervised objective function. This makes our model trainable for each testing sample and hence can adapt to its characteristics. 
In addition, previous TTT methods~\cite{sun2020test,liu2021ttt++,gandelsman2022test} are more computationally expensive than standard inference, as they update the model by computing the gradient on a batch of samples during the test time. The extra cost becomes important in the case of large models such as transformers. We investigate the simple dropout strategies on intermediate tokens to alleviate the issue. By randomly dropping part of tokens within a given image iteratively, we manage to construct a batch of different samples. This strategy not only improves the performance but also the efficiency.

\begin{figure}[t]
    \centering
    \includegraphics[width=\linewidth]{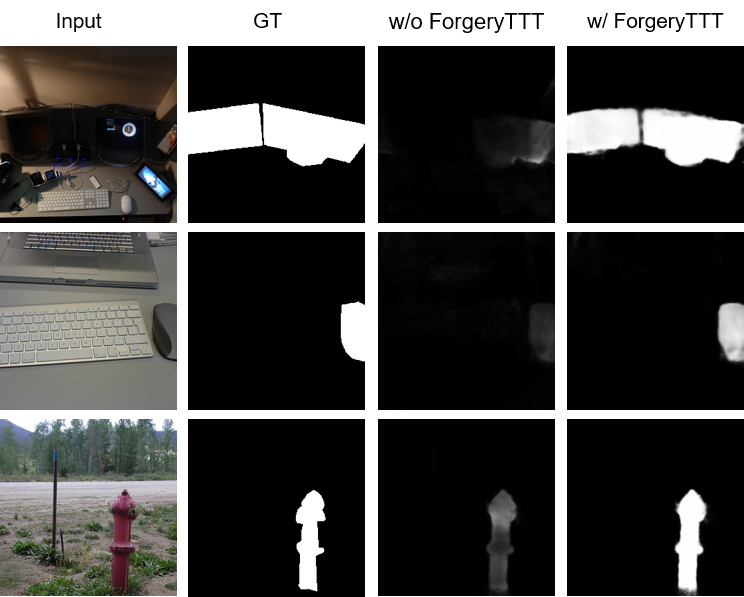}
    \caption{\textbf{Examples of localization results from our method, ForgeryTTT, on testing images.} Without ForgeryTTT, the model fails to accurately localize the forgery regions. However, performing adaptation on each image (\textit{w/ ForgeryTTT}), shows significantly better results.}
    \label{fig:brife}
\end{figure}

To evaluate our method in zero-shot image manipulation localization, we train the model on a large-scale synthetic dataset and evaluate it on five commonly used benchmarks.
The experimental results demonstrate that both the proposed auxiliary task and the test-time training strategy significantly boost the performance of image manipulation localization for unseen images and novel manipulation techniques, \textit{e.g.} an average improvement of 14.7$\%$ fixed threshold F1 score across various datasets.
The proposed one-to-batch sample generation test-time training strategy is both more effective and more efficient than the common strategy, leading to 1.8$\%$ better performance and 4.8$\times$ faster adaptation.
Our proposed method exhibits surprising gains, which finding the forgery region accurately through test-time training while the pre-trained model fails initially~(Figure~\ref{fig:brife}). 
When compared to state-of-the-art image manipulation localization methods in a zero-shot setting, our proposed method outperforms them by a substantial margin. 
Our approach also achieves superior performance even when compared to methods that incorporate the training split of the target dataset during training. 
Additionally, it surpasses other TTT approaches, underscoring the effectiveness of the proposed TTT strategy.


In summary, the contributions of this work are as follows:

\begin{itemize}


\item We introduce ForgeryTTT, to our best knowledge, the first test-time training framework specifically designed for zero-shot image manipulation localization.


\item We design a self-supervised task to enhance localization ability, demonstrating that it allows for more effective TTT than existing TTT approaches.

\item  Extensive experiments on five benchmarks demonstrate the superiority of ForgeryTTT, surpassing both zero-shot and non-zero-shot state-of-the-art image manipulation localization methods.

\end{itemize}



%% file: submit/related.tex

\begin{figure*}[t]
\centering
\includegraphics[width=\textwidth]{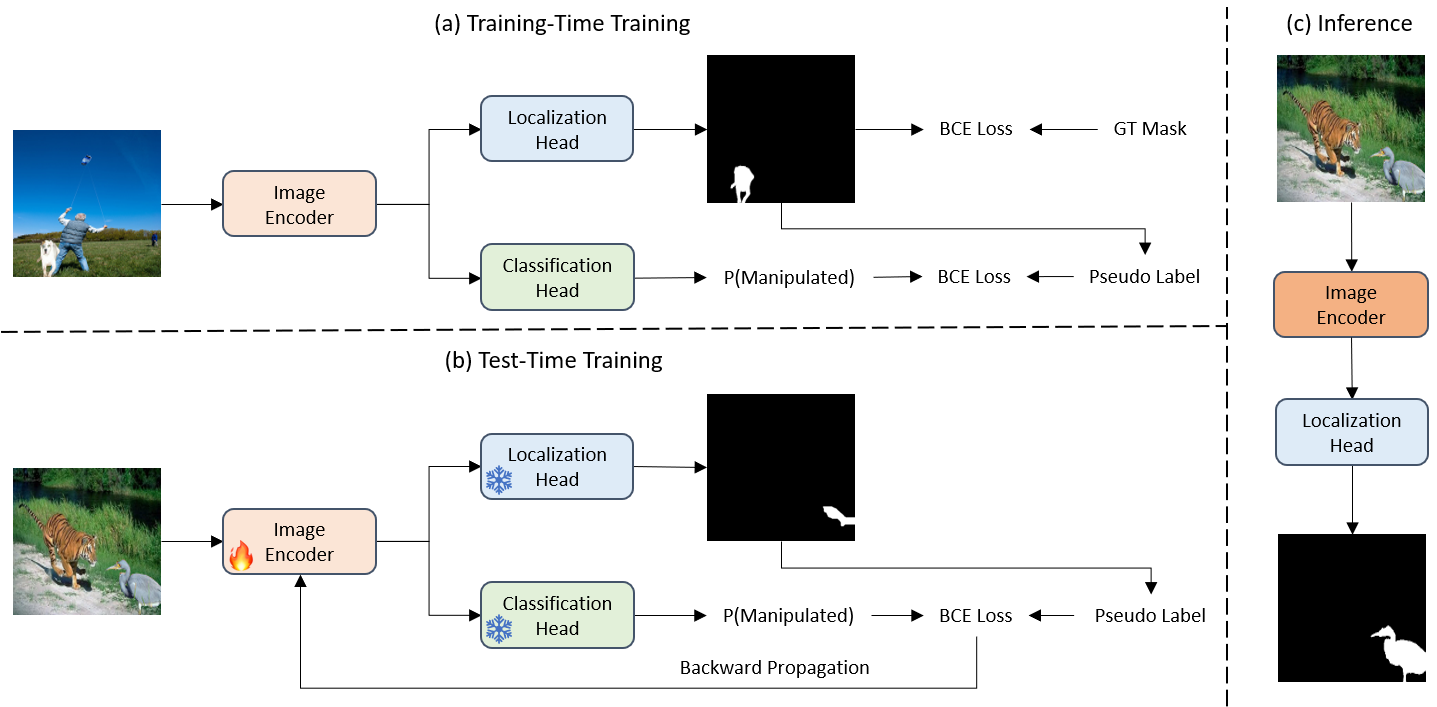}
\caption{\textbf{The overview of the proposed ForgeryTTT.} 
ForgeryTTT is a multi-task framework built upon the common encoder-decoder image manipulation localization network, which includes a shared image encoder, a localization head, and a classification head.
It is first learned to image manipulation localization and image manipulation classification on large-scale datasets.
Then, we employ a self-supervised loss based on the classification head to train the image encoder for each test image.
Finally, the updated model is used to localize the forgery region.
}
\label{fig:overview}
\end{figure*}

\section{Related Work}
\label{sec:related}

In this section, we present the work related to the proposed method from three aspects, including image manipulation localization, image manipulation classification, and test-time training.
\subsection{Image Manipulation Localization}
Early methods in image manipulation localization primarily rely on statistical analysis and handcrafted features.  They mainly focus on identifying jpeg compression~\cite{lin2009fast}, noise inconsistencies~\cite{noi}, and color filter array~\cite{cfa}, \textit{etc}. 
While effective in detecting global inconsistencies, these techniques often struggle with complex manipulations and precise localization.
Recent advancements in deep learning have significantly influenced the field. Deep neural networks now automatically learn discriminative forensic features from large-scale datasets, resulting in notable improvements~\cite{mantra,cun2018image,liu2021two,liu2023explicit,liu2023explicitv2, guillaro2023trufor,liu2024dh}. Most of these methods employ an encoder-decoder architecture, where an image encoder extracts the features and a decoder predicts the forgery mask.
To capture imperceptible clues beyond RGB images, some approaches transform the RGB image into the noise domain or frequency domain, providing complementary views for more effective forgery detection. ManTraNet~\cite{mantra} integrates the SRM filter~\cite{fridrich2012rich} and Bayar filter~\cite{bayar2018constrained} into the pre-trained backbone to extract features from both the noise map and the RGB image. 
Objectformer~\cite{wang2022objectformer} extracts RGB features and high-frequency features via two encoders and combines them as multimodal features.
Edge artifacts have also been utilized as crucial supervision signals to better capture details. 
MVSSNet~\cite{chen2021image} employs the Sobel filter to create a multi-level edge-supervised branch. 
TANet~\cite{shi2023transformer} refines the boundary by bilaterally exploring foreground and background cues.
However, the proliferation of fake images and the development of forgery techniques present ongoing challenges, while no one has yet focused on the robustness when confronted with unseen scenes.

\subsection{Image Manipulation Classification}
Traditional image manipulation classification methods initially extract hand-crafted features and then analyze statistical characteristics to differentiate the manipulated and authentic images. Fridrich~\etal~\cite{fridrich2003detection} proposed a copy-move forgery detection method based on approximate block matching. 
Pan~\etal~\cite{pan2010region} identified forgery images using a SIFT-based matching method. 
Shi et al. \cite{shi2007natural} employed multi-size block discrete cosine transform and Markov transition probabilities to detect splicing. 
In the deep learning era, image manipulation classification methods utilize deep neural networks to capture manipulation traces \cite{15,yu2023discrepancy,hua2023learning}. 
Zhang~\etal~\cite{zhang2016image} input image patch features into a five-layer neural network to determine whether the image has been manipulated. 
Bayar~\etal~\cite{15} proposed constrained convolutional neural networks that suppress the influence of image content on manipulation traces and adaptively extract manipulation features.
Recent works~\cite{mantra,psccnet} on image manipulation localization simultaneously address both image manipulation classification and localization. 
Different from those supervised methods, we design a mask-based self-supervised image manipulation classifier.

\begin{figure*}[t]
\centering
\includegraphics[width=\textwidth]{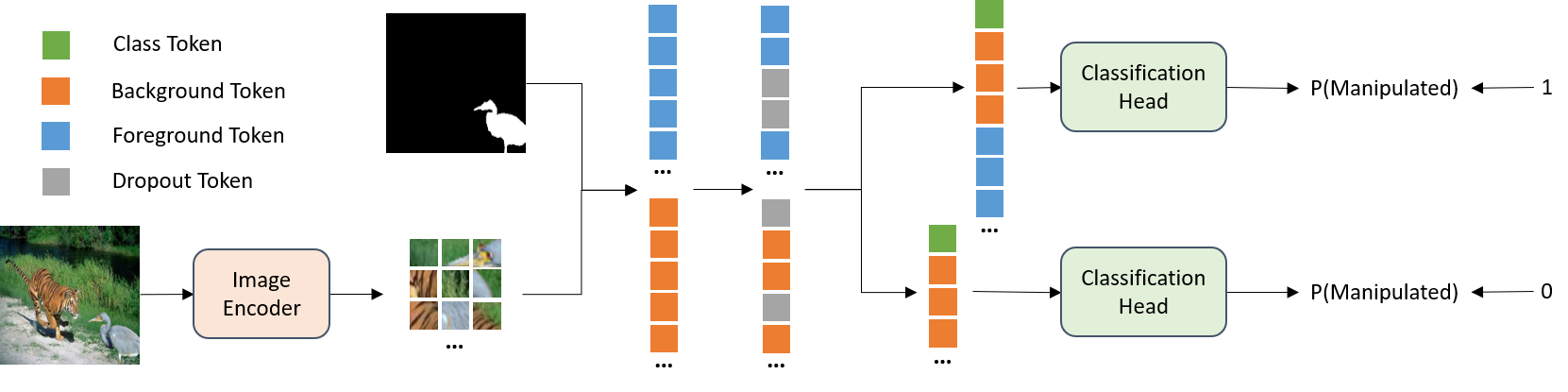}
\caption{\textbf{The proposed self-supervised image manipulation classification algorithm.} 
We first extract the image features using the image encoder.
Then, we group foreground manipulated tokens and background authentic tokens via a given mask.
Random token dropout is applied in both foreground and background tokens.
Next, the foreground tokens, background tokens, and class tokens are concatenated as manipulated queries, and background tokens and class tokens are concatenated as authentic queries.
Finally, the classification head is learned to distinguish these two kinds of queries.
}
\label{fig:ssl_pipeline}
\end{figure*}

\subsection{Test-Time Training}
The performance of deep neural networks may significantly drop in unseen test data that diverges from the training distribution. 
Test-time training~(TTT) aims at adapting the pre-trained model to out-of-distribution test data through self-supervised/ unsupervised object functions. 
Since the statistics in the BatchNorm layers represent domain-specific knowledge, the statistic-based methods recalculate the statistics to adapt the model to the test domain~\cite{schneider2020improving,wang2020tent,mirza2022norm}. Schneider~\etal~\cite{schneider2020improving} adjust the BatchNorm statistics of the model by estimating from the test sample.
Wang~\etal~\cite{wang2020tent} optimizes the affine parameters in the BatchNorm layers by minimizing the prediction entropy. 
Auxiliary head-based algorithms leverage the self-supervised learning task to adjust the features from the training domain to the test instance~\cite{sun2020test,gandelsman2022test,liu2021ttt++}. They typically consist of a shared image encoder, a primary decoder for the main task, and an auxiliary decoder for the self-supervised task.
Sun~\etal~\cite{sun2020test} develop an auxiliary head for rotation prediction and fine-tune the backbone using the auxiliary head during testing.
Gandelsman~\etal~\cite{gandelsman2022test} employ the mask autoencoder which reconstructs the test sample to update the pre-trained model.
Some works have explored effective TTT approaches in other areas~\cite{li2021test,sivaprasad2021uncertainty, liu2024depth}. It has not yet been explored in image manipulation localization and we take the first step.

%% file: submit/method.tex




\section{Method}
\label{sec:method}

To handle the evolving forgery images in the real world, we propose ForgeryTTT.
ForgeryTTT is composed of a shared image encoder, an image manipulation localization head, and a self-supervised image manipulation classification head.
We first train ForgeryTTT on a large-scale synthetic dataset. Then, we further train it for each test image via the self-supervised classification head and employ the updated model for inference.
The overall architecture of ForgeryTTT is shown in Figure~\ref{fig:overview}. 
In this section, we first introduce image manipulation localization and test-time training in section~\ref{sec:preliminaries}.
Then, we give the details of the proposed self-supervised image manipulation classification algorithm in section~\ref{sec:self_supervised iml}.
Finally, we introduce the two-stage pipeline of our method in section~\ref{sec:objective_function}.


\subsection{Preliminaries}
\label{sec:preliminaries}
\textbf{Image Manipulation Localization}.
Image manipulation localization involves localizing the pixel-level forgery mask in an RGB image. This is typically accomplished using an encoder-decoder network architecture~\cite{mantra,psccnet,liu2024dh}. In the image encoder, the image is first split into patch tokens, which are then processed through hierarchical transformer blocks to produce multi-scale representations.
These representations are subsequently fed into the localization head, which is responsible for predicting the forgery mask. The ground truth object mask is used to supervise the model via the binary cross-entropy loss.

\textbf{Test-Time Training.}
Test-time training (TTT) is introduced to enhance the generalization capability of models to out-of-distribution test data~\cite{sun2020test,wang2020tent,schneider2020improving}. A commonly employed TTT framework based on the auxiliary head~\cite{sun2020test} consists of a shared encoder $\encoder$, a main head $\demain$ for the main task, and an auxiliary head $\deauxiliary$ for self-supervised learning.
The TTT framework typically involves a two-stage training process. 
In the first stage, the network is trained using both the main loss $\lossmain$ and a self-supervised loss $\lossssl$:
\begin{equation} \label{eqn:1st_stage}
 \min _{\encoder, \demain, \deauxiliary} \lossmain + \sslweight \lossssl,
\end{equation}
where $\sslweight$ is the hyper-parameter to balance the two losses.

In the subsequent stage, which is known as test-time training (TTT), the encoder $\encoder$ is fine-tuned for each test sample based on the self-supervised objective function:
\begin{equation} \label{eqn:2st_stage}
 \min _{\encoder}\lossssl.
\end{equation}

This process allows the model to adapt to unseen data through the self-supervised learning task during testing, thereby improving its ability to generalize beyond the distribution of the training data.

\subsection{Self-supervised Image Manipulation Classification}
\label{sec:self_supervised iml}

Image manipulation classification has been integrated as an auxiliary task in several image manipulation localization algorithms~\cite{psccnet,li2024unionformer}. The classification head is trained in a supervised manner using labeled forgery and authentic images. 
The classification score helps detect global inconsistencies since forgery images exhibit a high response in at least one region, while authentic images show no response.
However, test-time training (TTT) necessitates a self-supervised auxiliary task to allow for model optimization without relying on labels during testing. To address this requirement, we develop a mask-based self-supervised image manipulation classification algorithm~(Figure~\ref{fig:ssl_pipeline}).

\begin{figure}[t]
\centering
\includegraphics[width=\linewidth]{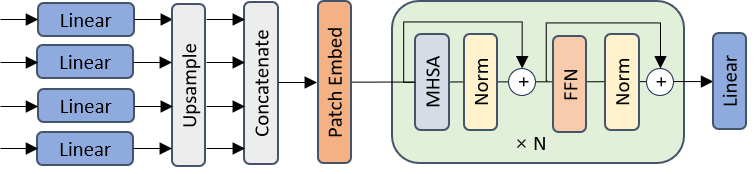}
\caption{\textbf{The details of the proposed classification head.}
The classification head merges the multi-scale features into tokens and outputs the probability of whether the given tokens are manipulated.
}
\label{fig:classification_head}
\end{figure}

\begin{figure*}[t]
\centering
\includegraphics[width=\textwidth]{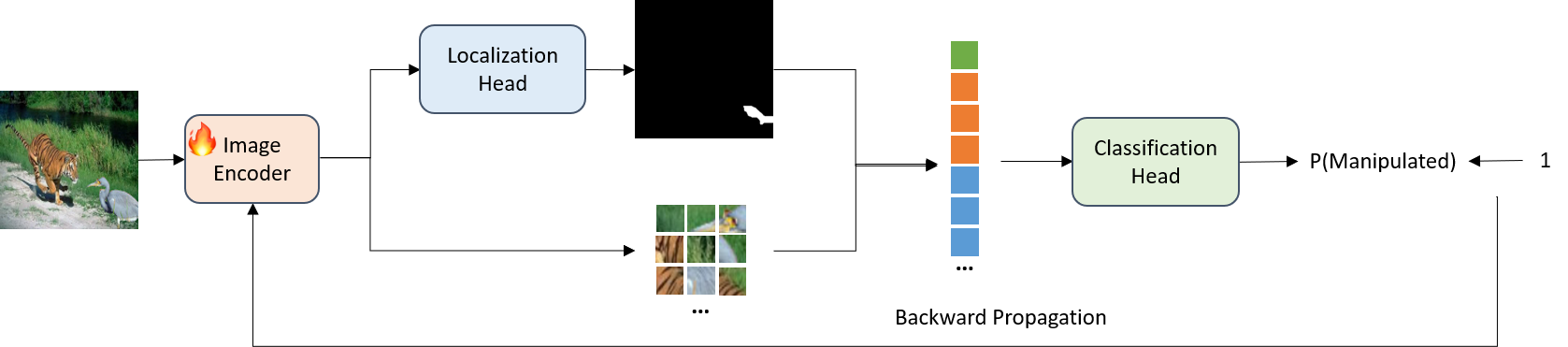}
\caption{\textbf{The details of the test-time training pipeline.}
For a given test image, we first extract its features via the image encoder and predict an initial mask via the localization head.
Then, we employ token group and token dropout on the features according to the predicted mask.
The remaining tokens are used to construct the manipulated query and serve as the pseudo-training sample to update the image encoder.
}
\label{fig:ttt_pipeline}
\end{figure*}

Given an input image $i \in \mathbb{R}^{h \times w \times 3}$, we extract multi-scale features $F=\left\{f_s: s \in (1,2,...,s_{max}) \right\}$ via the image encoder $\encoder$, where $f_s \in \mathbb{R}^{h_s \times w_s \times c_s}$ represent features extracted at scale $s$, and $h_s$, $w_s$, and $c_s$ are the height, width, and dimension of the feature at scale $s$, respectively.
The proposed classification head is shown in Figure~\ref{fig:classification_head}.
Specifically, we first merge the multi-scale features obtained by the image encoder into tokens. 
The multi-scale features are then unified in the channel dimension by the corresponding linear layer:
\begin{equation}
f_s=\operatorname{Linear_s}(f_s), s=1,2, \ldots, s_{max}.
\end{equation}

These features are upsampled to $h_1 \times w_1$ and concatenated together as $f$. The fused features $f$ are split into tokens with $p \times p$ patch size~(in contrast to the original resolution) via a patch embedding layer:
\begin{equation}
e=\operatorname{PatchEmbed}(f),
\end{equation}
where $e \in \mathbb{R}^{\frac{h}{p} \times \frac{w}{p} \times d}$.
Next, we select some tokens as query $q$.
The query and the learnable class token $c$ are fed into a series of transformer blocks:
\begin{equation} 
c_j, q_j  =L_j\left(\operatorname{Concat}[c_{j-1}, q_{j-1}]\right), j=1,2, \ldots, N, 
\end{equation}
where $L_j$ is the j-th transformer block. Each transformer block is composed of multi-head self-attention and feed-forward networks, together with layer normalization and residual connection.
Finally, we map the class token into the class probability via a linear layer:
\begin{equation} 
y =\operatorname{Linear}\left(c_N\right),
\end{equation}
where $y$ represents the probability of whether the given query is manipulated.

On top of the basic architecture, we further introduce several novel features, including label-free query construction and mask-based token dropout, which collectively make a powerful model.


\textbf{Lable-Free Query Construction.}
To adapt image manipulation classification to test-time training, we model it as a self-supervised task rather than a supervised one. The pseudo label is derived from the given mask. During training-time training, we employ the ground truth mask to group the foreground tokens and background tokens. 
The mask is downsampled using max-pooling to match the token size.
The foreground tokens and background tokens are used to construct the "manipulated" query, while the background tokens are used to construct the "authentic" query. These two types of tokens are concatenated with a class token to train the classifier in a self-supervised manner. 
During TTT, we construct only the manipulated query using the predicted mask to further train the image encoder with the pseudo-training samples. The manipulated query is reliable since the authentic query could be wrong when the predicted mask is inaccurate. 
Thus, we can use the constructed query to train a self-supervised classifier and fine-tune the pre-trained model with pseudo-training samples during testing.


\textbf{Mask-based Token Dropout.}
Previous methods that incorporate image manipulation classification as an auxiliary task input all tokens for classification. We believe it is redundant. A small number of authentic and manipulated tokens is sufficient to train a classifier. Inspired by recent advances in masked image modeling, we boost the performance via token dropout.
To achieve it, we first utilize a mask to differentiate between foreground-manipulated tokens and background-authentic tokens within a given set of image tokens. 
We then perform random dropout to select a subset of both foreground and background tokens, discarding the remaining tokens. This dropout process follows a uniform distribution, ensuring an equal representation of different types of tokens.
The dropout ratio \dropoutratio ~is the same in both foreground and background.
The benefit of this mask-based dropout strategy is straightforward. It boosts classification performance by eliminating redundant information since image manipulation classification does not rely on most of the tokens.

\subsection{Objective Function}
\label{sec:objective_function}
With the development of image editing technology, massive forgery images and advanced editing methods are challenging existing image manipulation localization algorithms. Mainstream methods often directly apply the models trained on large-scale datasets to the target test images, which makes them struggle to generalize to unseen images.  To overcome these limitations, we propose ForgeryTTT. ForgeryTTT first trains on a large-scale dataset and then adapts the pre-trained model to test samples through self-supervision during testing.

As illustrated in Figure~\ref{fig:overview}, 
ForgeryTTT is a multi-task framework composed of a shared image encoder~$\encoder$, an image manipulation localization head~$\demain$, and an image manipulation classification head~$\deauxiliary$.
The image encoder~$\encoder$ is a hierarchical transformer~\cite{liu2021swin} that produces a hierarchical representation for the input image~$\image$.
The localization head~$\demain$ is a lightweight multi-layer perception~\cite{xie2021segformer} that predicts the forgery mask.
The classification head~$\deauxiliary$ is composed of several transformer blocks and linear layers, which distinguish whether the given query is manipulated.

\subsubsection{Training-Time Training}
\label{sec:two_stage_part1}
During the first stage training-time training, ForgeryTTT simultaneously learns image manipulation localization and image manipulation classification. The ground truth mask~$\mask$ supervises the localization branch using the binary cross-entropy loss~$\lossbce$. The classification branch is trained in a self-supervised manner by sampling image tokens according to the ground truth mask to construct manipulated and authentic queries. These tokens are fed into the classifier to predict whether the given query is manipulated. 
The total objective function for this stage can be formulated as:
\begin{equation} \label{eqn:loss_training_time_training}
\lossbce(\demain(\enimage(\image)), \mask)  +   \sslweight \lossbce(\deauxiliary(\phi(\enimage(\image), \mask))),
\end{equation}
where $\phi$ denotes the whole process of token grouping, token dropout, and query construction.

\begin{figure}[t]
\centering
\includegraphics[width=\linewidth]{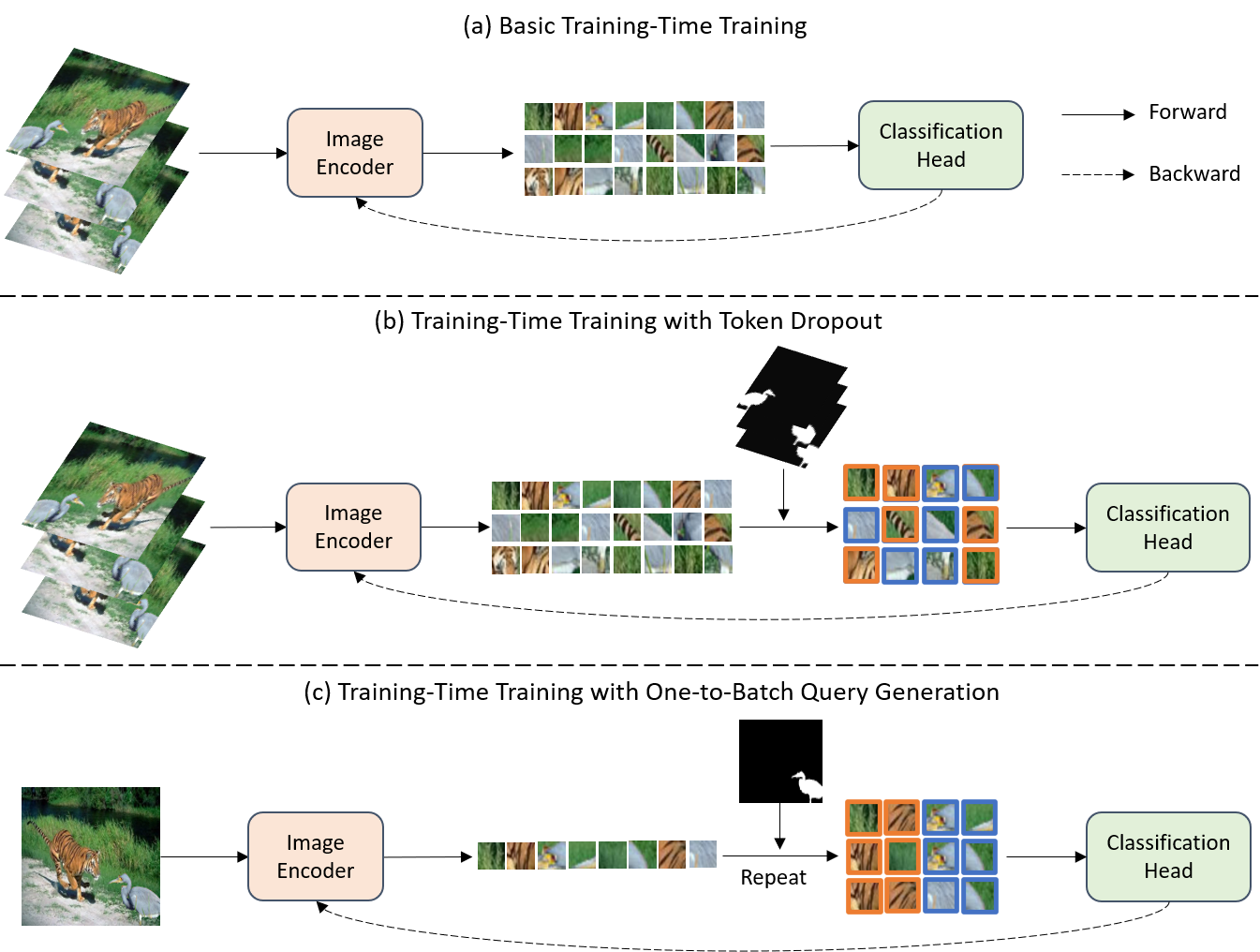}
\caption{\textbf{comparison of the different test-time training strategies.}
(a)~The basic test-time training strategy~(TTT-Base) uses a batch of images during the whole forward propagation.
(b)~Test-time training with token dropout~(TTT-TD) randomly drop some tokens since they are redundant for classification.
(c)~Test-time training with one-to-batch query generation~(TTT-OBQG) only encodes one image and constructs a batch of queries via repeating random dropout in the given tokens.
}
\label{fig:ttt_strategy}
\end{figure}

\subsubsection{Test-Time Training}
\label{sec:two_stage_part2}
In the second stage, which is known as test-time training, ForgeryTTT is required to generalize to unseen forgery images via optimizing the self-supervised objective function. As shown in Figure~\ref{fig:ttt_pipeline}, for each incoming test image~$\imagetest$, we first obtain its predicted mask~$\masktest$ using the image encoder~$\encoder$ and the localization head~$\demain$. We then randomly drop some tokens based on the predicted mask. These remaining tokens are fed into the classification head~$\deauxiliary$ as "manipulated" pseudo-training samples to fine-tune the pre-trained encoder~$\encoder$.

Leveraging the wonderful features of the aforementioned mask-based token dropout and label-free query construction, we further propose novel test-time training strategies~(Figure~\ref{fig:ttt_strategy}), named test-time token dropout~(TTT-TD) and one-to-batch query generation~(TTT-OBQG).

\textbf{Test-Time Token Dropout~(TTT-TD).}
Random token dropout during test-time training has been employed in methods such as TTT-MAE \cite{gandelsman2022test}. This technique randomly discards some tokens and uses masked image reconstruction as the self-supervised objective function.
The difference is that the key insight of our approach comes from the fact that image manipulation classification does not rely on most of the tokens~(as introduced in the previous section).
Consequently, we further incorporate token dropout into our test-time training process.
The dropout remains random, and the regions are specified using the predicted mask. 
We find it is also effective in our method.

\textbf{One-to-Batch Query Generation~(TTT-OBQG).}
TTT is generally performed on a batch of samples to derive more precise gradients. For instance, the basic TTT strategy utilizes data augmentation to acquire a batch of images. All images must pass through the image encoder, rendering TTT-Base inefficient in terms of running time and memory usage.
Here we propose one-to-batch query generation for efficient TTT.
Similarly, our strategy is built upon random dropout.
We randomly drop different tokens in the encoded image tokens to construct a batch of queries.
Therefore, we can construct a batch of queries with only one image being processed through the image encoder, which greatly reduces computation costs.

We also employ common multiple training steps to boost performance.
The model once fine-tuned on this sample serves as a more suitable initialization checkpoint for subsequent training compared to the pre-trained model. Consequently, the model undergoes an iterative update process, wherein each step fine-tunes the model based on the previous step. 
Overall, the k-th step optimization of ForgeryTTT can be formulated as:
\begin{equation} 
\label{eqn:multi_step_ttt}
\lossbce(\deauxiliary(\phi(\enimagei(\imagetest), \masktest))),
\end{equation}
We find that there is almost no difference if we replace $\masktest$ with $\masktesti$ in the above equation since our self-supervised task only needs pseudo-manipulated queries and a rough mask is sufficient.

\begin{figure*}[!h]
\centering
\includegraphics[width=0.95\textwidth]{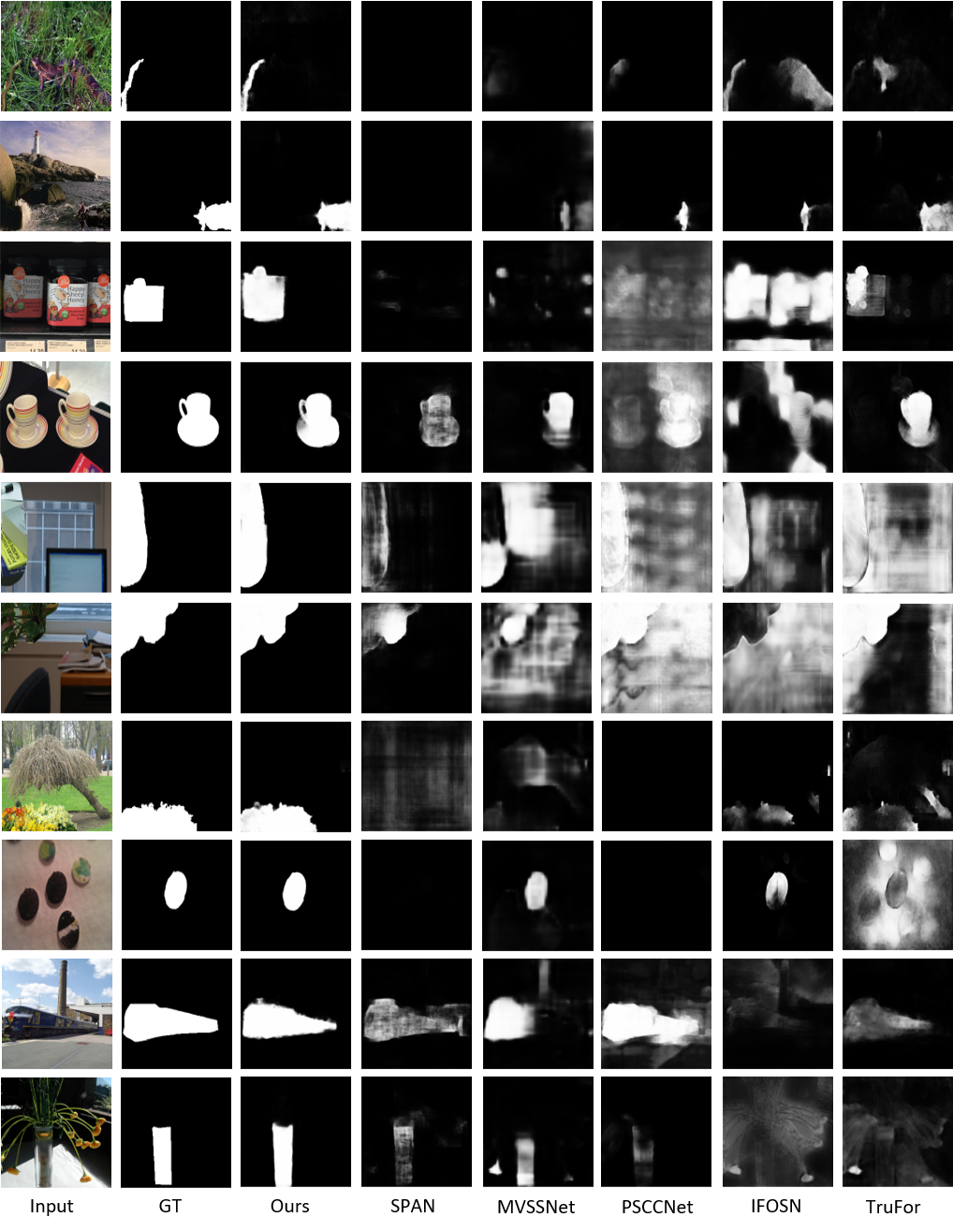}
\caption{\textbf{Qualitative comparison between the proposed method and existing SOTA methods.}
From top to bottom are the samples from the CASIA dataset~(1st and 2nd rows), Coverage dataset~(3rd and 4th rows), Columbia dataset~(5th and 6th rows), Nist16 dataset~(7th and 8th rows), and CocoGlide dataset~(9th and 10th rows).
}
\label{fig_sota}
\end{figure*}

\input{submit/table}

%% file: submit/table.tex
\newcommand{\fbest}{F$_{best}$}
\newcommand{\ffixed}{F$_{fix}$}
\newcommand{\auc}{AUC}
\newcommand{\acc}{ACC}

\newcommand{\sotatablestyle}[2]{\setlength{\tabcolsep}{#1}\renewcommand{\arraystretch}{#2}\centering\small}

\newcommand{\ablationtablestyle}[2]{\setlength{\tabcolsep}{#1}\renewcommand{\arraystretch}{#2}\centering\small}

\begin{table*}[t]
\caption{
Comparison with state-of-the-art image manipulation localization methods on CAISA~\cite{casia}, Coverage~\cite{coverage}, Colombia~\cite{columbia}, NIST16~\cite{nist16}, and CocoGlide~\cite{guillaro2023trufor} datasets. $^\dag$ means the training datasets have some overlap with the testing domain.}
\centering  
\sotatablestyle{2pt}{1.2}
 \resizebox{\textwidth}{!}
{
\begin{tabular}{l|llll|llll|llll|llll|llll|llll}
\toprule
 
 & \multicolumn{4}{c|}{CASIA} &
  \multicolumn{4}{c|}{Coverage} &
  \multicolumn{4}{c|}{Columbia} &
  \multicolumn{4}{c|}{NIST16}&
  \multicolumn{4}{c|}{CocoGlide}&
  \multicolumn{4}{c}{Average} \\ \cline{2-25}
 & \multicolumn{1}{c}{\fbest} &\multicolumn{1}{c}{\ffixed} & \multicolumn{1}{c}{\auc} &\multicolumn{1}{c|}{\acc} 
 & \multicolumn{1}{c}{\fbest} &\multicolumn{1}{c}{\ffixed} & \multicolumn{1}{c}{\auc} &\multicolumn{1}{c|}{\acc} 
 & \multicolumn{1}{c}{\fbest} &\multicolumn{1}{c}{\ffixed} & \multicolumn{1}{c}{\auc} &\multicolumn{1}{c|}{\acc} 
 & \multicolumn{1}{c}{\fbest} &\multicolumn{1}{c}{\ffixed} & \multicolumn{1}{c}{\auc} &\multicolumn{1}{c|}{\acc} 
 & \multicolumn{1}{c}{\fbest} &\multicolumn{1}{c}{\ffixed} & \multicolumn{1}{c}{\auc} &\multicolumn{1}{c|}{\acc} 
 & \multicolumn{1}{c}{\fbest} &\multicolumn{1}{c}{\ffixed} & \multicolumn{1}{c}{\auc} &\multicolumn{1}{c}{\acc}  \\
 \hline
ManTraNet~\cite{mantra}  & 18.0 & 32.0  & 64.4 &  50.0   & 48.6  & 31.7  & 76.0   &  50.0   & 65.0 &  50.8  & 81.0  & 50.0   & 22.5 & 17.2  & 62.4 & 50.0  & 67.3 & 51.6  & 77.8 & 50.0  & 44.3 & 36.7  & 72.3 & 50.0 \\
EXIF-SC ~\cite{ffn}  & 22.5 & 10.6  & 49.0 & 50.0   &  33.2 & 16.4  & 49.8 & 50.0   & 88.0 & 79.8  & 97.6 & 50.6   & 29.8 & 22.7  & 50.4 & 50.0  & 42.4 & 29.3  & 52.6 & 50.0  & 43.2 & 31.8  & 59.9 & 50.1 \\
SPAN$^\dag$~\cite{span}   & 16.9 & 11.2  & 48.0 & 48.7  & 42.8 & 23.5  & 67.0 & 60.5  & 87.3 & 75.9  & 99.9 & 95.1   & 36.3 & 22.8  & 63.2 & 59.7  & 35.0 & 29.8  & 47.5 & 49.1  & 43.7 & 32.6  & 65.1 & 62.6 \\
Noiseprint~\cite{cozzolino2019noiseprint} & 20.5 & 13.7  & 49.4 & -   & 34.2 & 22.9  & 52.5 & -   & 85.3 & 51.3  & 87.2 & -  & 34.5 & 19.6  & 61.8 & -  & 40.5 & 31.8  & 52.0 & -  & 43.0 & 27.9  & 60.6 & - \\
MVSSNet$^\dag$~\cite{chen2021image} & 65.0 & 52.8  & 93.2 & 80.0   & 65.9 & 51.4  & 73.7 & 54.5   & 78.1 & 72.9  & 98.4 & 66.7   & 37.2 & 32.0  & 57.9 & 53.8  & 64.2 & 48.6  & 65.4 & 53.6  & 62.1 & 51.5  & 77.6 & 61.9 \\
IFOSN~\cite{wu2022robust} & 67.6 & 55.3  & 73.5 & 63.5   & 47.2 & 30.4  & 55.7 & 51.0   & 83.6 & 75.3  & 88.2 & 52.2  & 44.9 & 33.0  & 65.8 & 55.3  & 58.9 & 42.8  & 61.1 & 56.7  & 60.4 & 47.4  & 68.9 & 55.7 \\
CATNetv2$^\dag$~\cite{kwon2022learning} & 85.2 & 75.2  & 94.2 & 83.8   & 58.2 & 38.1  & 68.0 & 63.5   & 92.3 & 85.9  & 97.7 & 80.3   & 41.7 & 30.8  & 75.0 & 59.7  & 60.3 & 43.4  & 66.7 & 58.0  & 67.5 & 54.7  & 80.3 & 69.1 \\
PSCCNet$^\dag$~\cite{liu2022pscc} & 67.0 & 52.0  & 86.9 & 68.3   & 61.5 & 47.3  & 65.7 & 55.0   & 76.0 & 60.4  & 30.0 & 50.8   & 21.0 & 11.3  & 48.5 & 45.6  & 68.5 & 51.5  & 77.7 & 66.1  & 58.8 & 44.5  & 61.8 & 57.2 \\
TruFor$^\dag$~\cite{guillaro2023trufor} & 82.2 & 73.7  & 91.6 & 81.3   & \bf73.5 & 60.0  & 77.0 & 68.0   & 91.4 & 85.9  & 99.6 & \bf98.4   & 47.0 & 39.9  & 76.0 & 66.2  & 72.0 & 52.3  & 75.2 & 63.9  & 73.2 & 62.4  & 83.9 & 75.6 \\
UnionFormer$^\dag$~\cite{li2024unionformer} & \bf86.3 & \bf76.0  & \bf95.1 & 84.3   & 72.0 & 59.2  & 78.3 & 69.4   & 92.5 & 86.1  & \bf99.8 & 97.9   & 48.9 & 41.3  & \bf79.3 & 68.0  & 74.2 & 53.6  & 79.7 & 68.2  & 74.8 & 63.2  & \bf86.4 & 77.6 \\
Ours & 80.4 & 72.2  & 91.6 & \bf87.1   & 70.8 & \bf63.0  & \bf87.1 & \bf77.2   & \bf95.7 & \bf92.7  & 87.9 & 87.4   & \bf55.8 & \bf44.3  & 76.0 & \bf71.7  & \bf76.8 & \bf65.4  & \bf86.6 & \bf72.9  & \bf75.9 & \bf67.5 & 85.8 & \bf79.3 \\
 \bottomrule
\end{tabular}
}
\label{tab:sota_iml}
\end{table*}

\begin{table}[t]
\centering
\caption{Comparison with state-of-the-art test-time training methods on CAISA~\cite{casia}, Coverage~\cite{coverage}, Colombia~\cite{columbia}, NIST16~\cite{nist16}, and CocoGlide~\cite{guillaro2023trufor} datasets using \ffixed.}
\ablationtablestyle{2pt}{1.2}
\begin{tabular}{l|ccccc}
\toprule
    & CASIA & Coverage & Columbia & NIST16 & CocoGlide \\ \hline
  Baseline & 61.4 & 35.6 & 89.4 & 34.7 & 43.0 \\
  BN~\cite{schneider2020improving} &  62.5 &   38.4 & 89.1 &  35.3 & 44.1    \\
 TENT~\cite{wang2020tent} &   62.1  & 37.7 &  88.6 & 34.4 & 44.6  \\ 
 TTT-ROT~\cite{sun2020test} &  62.7 & 40.1 &  90.2 & 36.8 & 45.7 \\
 TTT-MAE~\cite{gandelsman2022test} &  63.0 &  39.4 &   89.9& 35.1 & 45.2 \\
 Ours &  \bf72.2 &  \bf63.0 &  \bf92.7 & \bf44.3 & \bf65.4 \\
\bottomrule
\end{tabular}
\label{tab:sota_ttt}
\end{table}

%% file: submit/exp.tex
\section{Experiments}
\label{sec:exp}

This section shows the experimental results for image manipulation localization of our method with other state-of-the-art methods.
Firstly, we illustrate the dataset and implementation details in section~\ref{sec:Datasets} and section~\ref{sec:Implementation Details}, respectively.
Then, we compare the proposed method with other state-of-the-art methods on several benchmarks in section~\ref{sec:comparison}. 
Next, we give a detailed analysis of each component of the proposed method in section~\ref{sec:ablation}.
Finally, we evaluate its robustness in several distortion settings in section~\ref{sec:robustness}.

\subsection{Datasets}
\label{sec:Datasets}



We utilize multiple datasets, including SynCOCO, CASIA~\cite{casia}, Coverage~\cite{coverage}, Columbia~\cite{columbia}, NIST16~\cite{nist16}, and CocoGlide~\cite{guillaro2023trufor}, in our experiments.

\textbf{Training Data.} 
We create a synthetic dataset named SynCOCO to train our model. This dataset is built upon MSCOCO~\cite{mscoco} and contains forgery images of several manipulation types, including splicing, copy-move, and removal. For splicing, we extract an annotated instance from a random image and paste it into another image. For copy-move, we extract an annotated instance from a random image and paste it into a different region within the same image. For removal, we delete an annotated instance from an image and employ a cutting-edge image inpainting algorithm~\cite{suvorov2022resolution} to restore the region. In total, the synthetic dataset comprises 150,000 labeled forgery images.

\textbf{Testing Data.} 
We evaluate our method on five publicly available benchmarks. 
CASIA~\cite{casia} is composed of CASIA v1~(5,123 images) and CASIA v2~(921 images), including splicing and copy-move samples. We evaluate our method on CASIA v2. 
Coverage~\cite{coverage} is a copy-move forgery dataset containing 100 manipulated images. We use the test split (25 images) to evaluate the proposed method. 
Columbia~\cite{columbia} includes 180 forgery images designated for splicing detection. We evaluate our method on the entire dataset. 
NIST16~\cite{nist16} contains 564 forgery images involving splicing, copy-move, and removal types. We use the test split (160 images) to evaluate the proposed method. 
CocoGlide~\cite{guillaro2023trufor} comprises 512 images generated from the COCO 2017 validation set using the GLIDE diffusion model~\cite{nichol2022glide}. We evaluate our method on the entire dataset.



\subsection{Implementation Details}
\label{sec:Implementation Details}
We implement our algorithm using Pytorch framework and all the experiments are performed on a single NVIDIA A40 GPU. As for optimization, we use the Adam~\cite{adam} optimizer.
The initial learning rate is set to $2e^{-4}$, which is adjusted by the exponential decay strategy with a decay rate of 0.9.
We employ a fixed learning rate of $2e^{-5}$ during TTT.
The models are trained for 20 epochs on SynCOCO dataset and 10 epochs during TTT on each test sample. 
The mini-batch size is equal to 32. 
The input image is resized into a fixed resolution of $512 \times 512$. Random horizontal flipping, resizing, and cropping are used for data augmentation.

As for the network, we adopt Swin-Tiny~\cite{liu2021swin} as the image encoder, the lightweight ALL-MLP decoder~\cite{xie2021segformer} as the localization head, and the number of transformer blocks in the classification head is set to 5. 
The loss weight $\lambda$ in training-time training~(equation \ref{eqn:loss_training_time_training}) is set to 0.01.
The dropout ratio for token dropout is 0.5 and the patch size in the classification head is 16 $\times$ 16~(section \ref{sec:self_supervised iml}).

We evaluate the proposed method in image manipulation localization using several metrics following~\cite{guillaro2023trufor,li2024unionformer}.
We report the F1 score using both the best threshold~(\fbest) and the default 0.5 threshold~(\ffixed) to measure pixel-level performance.
For image-level analysis, we utilize the Area Under the Curve~(\auc) and balanced accuracy metrics~(\acc). 
All these metrics are the larger the better.


\begin{table}[t]
\centering
\caption{Ablation study of the self-supervised image manipulation classification algorithm on CAISA~\cite{casia}, Coverage~\cite{coverage}, Colombia~\cite{columbia}, NIST16~\cite{nist16}, and CocoGlide~\cite{guillaro2023trufor} datasets using \ffixed.}
\ablationtablestyle{3pt}{1.2}
\begin{tabular}{l|ccccc}
\toprule
 Self-supervised Head &  - &  \checkmark & \checkmark &  \checkmark &   \checkmark \\
 Token Dropout & - &  - & - &  \checkmark &   \checkmark \\
 Test-Time Training & -  & -  & \checkmark & - &  \checkmark \\ \hline
 CASIA &  61.4 & 63.1  & 66.8 &  65.6 & \bf72.2  \\
 Coverage & 35.6  & 35.8  &  60.2  &  37.3 &  \bf63.0 \\
 Columbia & 89.4  &  88.9 &  90.5 & 90.2 & \bf92.7 \\
 NIST16 &  34.7 & 35.4  &\bf 47.1  & 32.9 & 44.3 \\
 CocoGlide   &  43.0   & 44.2  & 64.0  & 46.7 &  \bf65.4 \\ 
\bottomrule
\end{tabular}
\label{tab:ablation_arch}
\end{table}

\begin{table}[t]
\centering
\caption{Ablation study of different queries in test-time training on CAISA~\cite{casia}, Coverage~\cite{coverage}, Colombia~\cite{columbia}, NIST16~\cite{nist16}, and CocoGlide~\cite{guillaro2023trufor} datasets using \ffixed.}
\ablationtablestyle{3pt}{1.2}
\begin{tabular}{l|ccccc}
\toprule
  & CASIA & Coverage & Columbia & NIST16 & CocoGlide \\ \hline
 - & 65.6 & 37.3 & 90.2 & 32.9 & 46.7   \\
 Mani. Query & \bf72.2 & \bf63.0 & \bf92.7 & \bf44.3 & \bf65.4   \\
 Auth. Query & 47.5 & 29.1 &  77.8 & 18.8 & 36.5  \\ 
  Both Queries & 54.7 &  35.7 & 81.0 & 20.9 & 40.2 \\

\bottomrule
\end{tabular}
\label{tab:ablation_loss}
\end{table}

\begin{table}[t]
\centering
\caption{Ablation study of the patch size on CAISA~\cite{casia}, Coverage~\cite{coverage}, Colombia~\cite{columbia}, NIST16~\cite{nist16}, and CocoGlide~\cite{guillaro2023trufor} datasets using \ffixed.}
\ablationtablestyle{3pt}{1.2}
\begin{tabular}{l|ccccc}
\toprule
    & CASIA & Coverage & Columbia & NIST16 & CocoGlide \\ \hline
 16 $\times$ 16 & \bf72.2 & \bf63.0 & \bf92.7 & \bf44.3 & 65.4    \\
 32 $\times$ 32  &   71.5 &  56.1 &  91.6 & 43.1 & \bf69.6  \\ 
 64 $\times$ 64 &  70.3 &  52.7 &   90.4& 42.4 & 68.3 \\
\bottomrule
\end{tabular}
\label{tab:ablation_patch}
\end{table}

\begin{table}[t]
\centering
\caption{Ablation study of the dropout ratio on CAISA~\cite{casia}, Coverage~\cite{coverage}, Colombia~\cite{columbia}, NIST16~\cite{nist16}, and CocoGlide~\cite{guillaro2023trufor} datasets using \ffixed.}
\ablationtablestyle{3pt}{1.2}
\begin{tabular}{l|ccccc}
\toprule
    & CASIA & Coverage & Columbia & NIST16 & CocoGlide \\ \hline
0.1 &  70.6 &   58.8 &  90.8 &   44.1 & 63.0    \\
0.3  &   71.4 &  60.4 &  91.6 & \bf44.9 & 64.2  \\ 
0.5 & \bf72.2 & \bf63.0 & \bf92.7 & 44.3 & \bf65.4 \\ 
0.7 &  71.8 &  61.7 &   92.2 & 42.7 & 65.0 \\ 
0.9 &  69.2 &  59.5 &  91.0 & 41.8 & 63.7 \\
\bottomrule
\end{tabular}
\label{tab:ablation_dropout_rtio}
\end{table}

\begin{table}[t]
\centering
\caption{Ablation study of different test-time training strategies.}
\ablationtablestyle{3pt}{1.2}
\begin{tabular}{l|ccc}
\toprule
    & \ffixed & Memory~(mb) & Runtime~(ms) \\ \hline
TTT-Base & 65.7 & 43,600 & 1,250  \\
TTT-TD & \bf67.8 & 35,600 &  1,010  \\
TTT-OBQG & 67.5 & \bf7,100 &  \bf260  \\
\bottomrule
\end{tabular}
\label{tab:ablation_strategy}
\end{table}

\subsection{Comparison with state-of-the-art methods}
\label{sec:comparison}

We compare the performance of our method with other state-of-the-art image manipulation methods~\cite{mantra,span,ffn,cozzolino2019noiseprint,chen2021image,wu2022robust,kwon2022learning,liu2022pscc,guillaro2023trufor}.
Since our method is designed for zero-shot image manipulation localization, the model is only trained on a large-scale synthetic dataset and evaluated directly on the test split of the target datasets. Instead, some methods further fine-tune their models on the training split of the target dataset~\cite{span,psccnet} or include the training split into their training data~\cite{guillaro2023trufor,li2024unionformer}. 
For a fair comparison, we divided them into two categories by different training protocols:
(1) None zero-shot methods: SPAN~\cite{span}, MVSSNet~\cite{chen2021image}, CATNetv2~\cite{kwon2022learning}, PSCCNet~\cite{liu2022pscc}, TruFor~\cite{guillaro2023trufor}, and UnionFormer~\cite{li2024unionformer}.
(2) Zero-shot methods: ManTraNet~\cite{mantra}, EXIF-SC~\cite{ffn}, Noiseprint~\cite{cozzolino2019noiseprint}, IFOSN~\cite{wu2022robust}, and our ForgeryTTT.

Table~\ref{tab:sota_iml} describes the comparison of \fbest, \ffixed, \auc, and \acc ~between our method and other state-of-the-art methods. 
Specifically, our method is much better than previous zero-shot methods~(ManTraNet~\cite{mantra}, EXIF-SC~\cite{ffn}, Noiseprint~\cite{cozzolino2019noiseprint}, and IFOSN~\cite{wu2022robust}). 
Our approach outperforms them on all five benchmarks and achieves an~\ffixed~improvement of 20.1$\%$ on average at least.
On the other hand, our approach also achieves the top performance when compared to the methods that include the training split of the target dataset into the training dataset~(CATNetv2~\cite{kwon2022learning}, TruFor~\cite{guillaro2023trufor}, and UnionFormer~\cite{li2024unionformer}) or fine-tuning the pre-trained model on the training split~(SPAN~\cite{span}, MVSSNet~\cite{chen2021image}, and PSCCNet~\cite{liu2022pscc}).
Our method gets the highest average ~\ffixed~, even though our model has not seen these images before. 
It is worth noting that our approach performs best on the CocoGlide dataset, which is built with a prompt-based generative artificial intelligence technique~\cite{nichol2022glide}. None of these methods have included this kind of data during training. This shows that our approach has strong potential to be generalized to unseen scenarios.
We show the visual comparison of our method and other state-of-the-art methods in Figure~\ref{fig_sota}. The proposed method performs better than others on five benchmarks, which proves the superiority of the proposed method.

We further analyze the efficiency of the proposed method.
The parameters of each component in our model are 27.5M for the image encoder, 0.6M for the localization head, and 5.1M for the classification head. Consequently, we introduce only a small number of additional parameters to the baseline model. 
As for comparison, our model~(33.2M) is smaller than other top competitors, such as TruFor~(68.7M), CATNetv2~(114.3M),  IFOSN~(128.8M), and MVSSNet~(146.9M).
Our model takes about 30 hours on one A40 GPU for training, while TruFor takes more than 14 days for training on one A6000 GPU.
Our full model takes 12 milliseconds per frame for inference and 260 milliseconds per frame for TTT.
As an inherent drawback of TTT, it does bring additional computational time.
Therefore we also try to mitigate its impact. Equipped with the proposed strategy, the proposed method becomes more effective and faster than the basic strategy~(please refer to the next section).

We also compare the proposed method with state-of-the-art TTT methods~\cite{schneider2020improving,wang2020tent,sun2020test,gandelsman2022test}.
For statistics-based TTT methods~\cite{schneider2020improving,wang2020tent}, we directly apply TTT to the baseline model, and we follow a two-stage training pipeline for the TTT methods with an auxiliary head~\cite{sun2020test,gandelsman2022test}. 
All these methods are pre-trained on SynCOCO and evaluated on the test split of five benchmarks.
As shown in Table~\ref{tab:sota_ttt}, the proposed method outperforms all other TTT methods significantly.
On the one hand, although other TTT methods work well in some corruptions such as fog, snow, and rain, they may be less effective against natural domain shifts.
On the other hand, designing an auxiliary task that is closely related to the main task has a significant impact on the performance of TTT. The proposed ForgeryTTT is proved to be advantageous to image manipulation localization, which brings significant improvements when applied to unseen forgery images.

\subsection{Ablation Analysis}
\label{sec:ablation}

\begin{figure}[t]
\centering
\includegraphics[width=\linewidth]{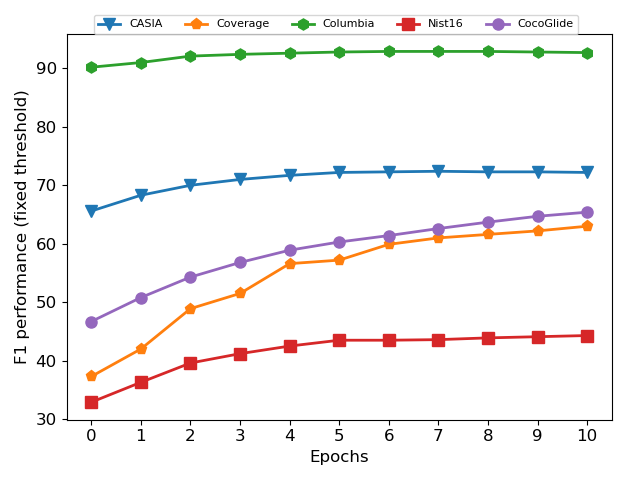}
\caption{\textbf{Performance comparison in different TTT epochs on CAISA~\cite{casia}, Coverage~\cite{coverage}, Colombia~\cite{columbia}, NIST16~\cite{nist16}, and CocoGlide~\cite{guillaro2023trufor} datasets.}
The performance is getting better as TTT goes on. Different datasets have different optimal epochs to achieve the best performance.
}
\label{fig_epoch}
\end{figure}

\begin{figure*}[t]
\centering
\includegraphics[width=\textwidth]{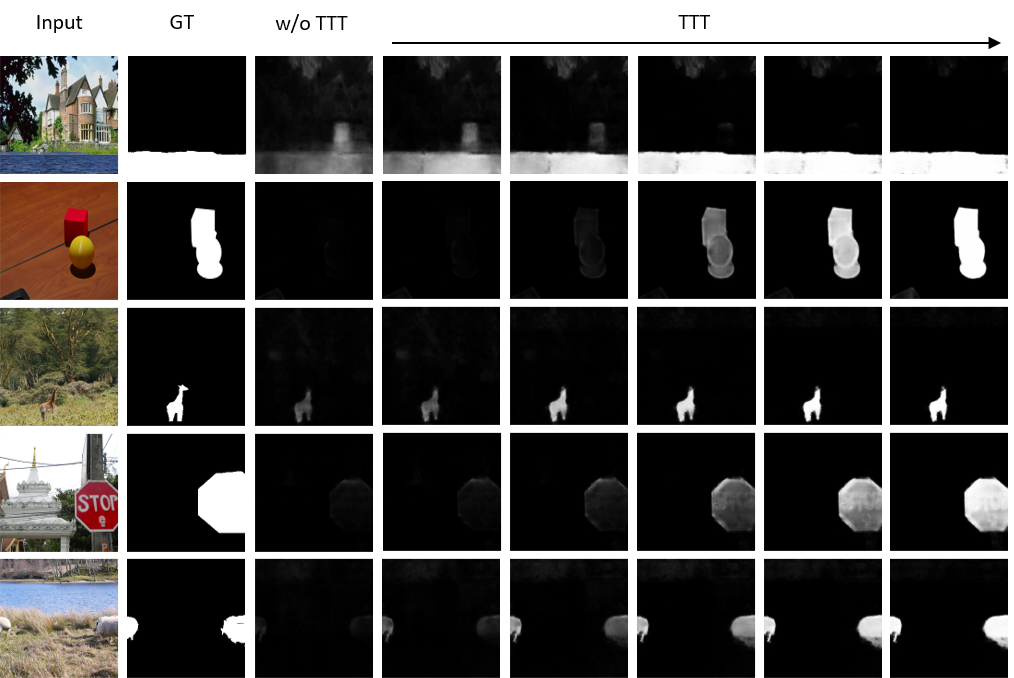}
\caption{\textbf{Qualitative results of in different TTT epochs.}
Our method shows less effectiveness before TTT when confronted with unseen forgery images.
The results are progressively improved during TTT, and our method finally accurately localizes the forged regions.
}
\label{fig_ttt}
\end{figure*}

\paragraph{Architecture Design}
We first verify the proposed architecture and the results are shown in Table~\ref{tab:ablation_arch}.
The baseline is set to the common encoder-decoder model. 
First, the proposed self-supervised image manipulation classification head is beneficial to image manipulation localization in training-time training and obtains significant gains during test-time training, which indicates that our designed self-supervised image manipulation classification algorithm suits image manipulation localization very well.
Second, the model with token dropout has a better performance in both training-time training and test-time training, which shows that reducing redundancy is effective for image manipulation classification.
The improvements are stable in different datasets, demonstrating the effectiveness in handling unseen images and unseen manipulation techniques.

We then explore constructing different queries in test-time training. As shown in Table~\ref{tab:ablation_loss}, when we try to use the authentic query or both authentic and manipulated queries, the performance will be much worse than only using the manipulated query for TTT and even not performing TTT. This is because the predicted mask is inaccurate therefore the constructed authentic query may be wrong and harm test-time training.

Next, we validate the influence of the patch size of image tokens. As shown in Table~\ref{tab:ablation_patch}, the performance will drop as the patch size increases in most datasets.
On the one hand, the number of patches available for sampling decreases, and the diversity of queries constructed decreases.
On the other hand, a larger patch size means less detailed information, which may harm image manipulation classification.

Finally, We find that the dropout ratio is also important since it affects the random dropout process.
The results of different dropout ratios are presented in Table~\ref{tab:ablation_dropout_rtio}.
The model will degenerate into a simple image manipulation classification model when the dropout ratio is set to 0, and the performance is poor when the dropout ratio is too low~($<$0.3) or too high~($>$0.7). We take 0.5 as the default dropout ratio since it works well in most datasets.
As a result, dropout is important in constructing different queries, an appropriate dropout ratio is beneficial in sampling enough tokens for classification.

\paragraph{TTT Scheme} 
We present three test-time training schemes in section~\ref{sec:objective_function}, including basic test-time training~(TTT-Base), test-time training with token dropout~(TTT-TD), and test-time training with one-to-batch sample generation~(TTT-OBSG). We compare them in Table~\ref{tab:ablation_strategy}.
A batch of samples enables TTT to estimate more accurate gradients but at the cost of increased GPU memory usage and runtime. 
For instance, when we use only one image in TTT-Base, the average \ffixed ~is 61.9, and it rises to 65.7 if we feed 32 samples obtained by data augmentation. However, this comes at the cost of increasing GPU memory usage from 4,300 mb to 43,600 mb and runtime from 24 milliseconds to 1,250 milliseconds.
Next, the average \ffixed ~improves to 67.8 as we adopt TTT-TD, and the GPU memory usage and runtime decrease to 35,600 mb and 1,010 milliseconds, respectively. This shows that token dropout is also effective for test-time training because using all tokens for classification is redundant.
Finally, when we adopt TTT-OBSG, the GPU memory usage and runtime drop rapidly to 7,100 mb and 260 milliseconds, respectively. The average \ffixed ~lost only 0.3 compared to TTT-TD and still 1.8 better than TTT-Base, which demonstrates the effectiveness and efficiency of the proposed TTT-OBSG.

We show the results of TTT epochs on performance in Figure~\ref{fig_epoch}.
Specifically, remarkable improvements can be observed across all datasets at the first few TTT epochs. As the number of epochs increases, the performances on some datasets begin to saturate. For convenience, we used 10 epochs as default on all datasets since more epochs could not bring noticeable improvements. 
We also plot the results obtained by our method in Figure~\ref{fig_ttt}. 
Our model typically can not locate the forgery region initially for the given unseen forgery images. As TTT goes on, we observe that the prediction becomes more and more accurate.

\begin{table}[t]
\centering
\caption{Robustness analysis under several distortion settings on CAISA~\cite{casia} dataset using $F_{fixed}$.}
\ablationtablestyle{3pt}{1.2}
\begin{tabular}{l|cccc}
\toprule
   & MVSSNet & IFOSN & TruFor & Ours \\ \hline
None   &  52.8 & 55.3 & 73.7  &  72.2       \\ \hline
GaussianBlur(k=3)   &  \bf-12.8 & -14.6 &  -31.3  &   -24.2    \\
GaussianBlur(k=5)  & \bf-18.2  & -19.5  & -34.5  & -28.5     \\ \hline
GaussianNoise($\sigma$=3)   & -9.3 & -3.5  &  -2.9 &  \bf-1.6   \\
GaussianNoise($\sigma$=5)   &  -20.0 & -3.8  &  -4.6 &  \bf-2.1   \\\hline
JPEGCompress(50)   & -37.7 & -26.9  &  \bf-14.2  & -20.4  \\
JPEGCompress(100)   & -21.0 &  -8.0 &  \bf-3.4  & -3.6    \\ \hline
Transmission(Facebook)   & -5.9 & -4.0 & -2.1  & \bf-1.9    \\
Transmission(Whatsapp)   & -8.4 & -2.9 &  -2.4 &  \bf-1.0   \\ 
Transmission(Weibo)   & -4.8 & -4.6 & -6.1   &  \bf-1.5      \\
Transmission(Wechat)   & -18.9 & -9.9 &  -12.2  &  \bf-9.0   \\ 
\bottomrule
\end{tabular}
\label{tab:robustness}
\end{table}

\subsection{Robustness Analysis}
\label{sec:robustness}
We evaluate the robustness under several distortion settings. Table~\ref{tab:robustness} shows the results on CAISA datasets of MVSSNet, IFOSN, TruFor, and our method, on which Gaussian Blur, Gaussian Noise, JPEG compression, and online social network transmission process the images. Specifically, the proposed method shows more stable results than others in most cases, which gains the least performance drop under various attacks.
Our method performs poorly when faced with highly compressed and Gaussian blurred images, and we notice that the performance of TruFor also drops significantly in these cases. Instead, MVSSNet and IFOSN perform better on Gaussian blurred images.
We think this is because of the different kinds of backbones.   Our method and TruFor use a transformer-based backbone, while MVSSNet and IFOSN use a CNN-based backbone. When image quality is severely degraded, the non-semantic discrepancies between manipulated and authentic regions become smaller, while local inconsistencies in manipulated boundaries are more important.
In addition, we believe that including these types of data augmentation in training could help improve the robustness of these attacks.

%% file: submit/conclusion.tex
\section{Conclusion}
\label{sec:conclusion}
In this work, we propose a novel image manipulation localization method termed ForgeryTTT. The pre-trained model effectively generalizes to unseen forgery images by fine-tuning the model for each test sample at test time.
We present a multi-task framework that simultaneously performs image manipulation localization and image manipulation classification, in which the classification head is learned in a self-supervised manner.
For each test sample, we first fine-tune the model with the self-supervised objective function, and then make a better prediction using the updated model.
We also explore some well-designed strategies to further enhance our model.
Extensive experiments in five publicly available image manipulation localization benchmarks demonstrate significant improvements in handling unseen forgery images, as well as outperform the existing image manipulation localization methods.
Our approach is promising for its excellent zero-shot performance, which shows its potential in generalizing the model to unseen forgery images in the real world.
In the future, we will develop a more powerful zero-shot forensics tool that covers other modalities to combat various forms of fake content on the internet.
We hope that this work can provide new ideas for other research on multimedia forensics.